# Approximate N-Gram Markov Model for Natural Language Generation

### Hsin-Hsi Chen    and    Yue-Shi Lee

Department of Computer Science and Information Engineering
National Taiwan University
Taipei, Taiwan, R.O.C.
E-mail: hh_chen@csie.ntu.edu.tw

### Abstract
This paper proposes an Approximate n-gram Markov Model for bag generation. Directed word association pairs with distances are used to approximate (n-1)-gram and n-gram training tables. This model has parameters of word association model, and merits of both word association model and Markov Model. The training knowledge for bag generation can be also applied to lexical selection in machine translation design.

## 1. Introduction

Natural language generation (Zock and Sabah, 1988; Dale, Mellish and Zock, 1990) forms an important component of many natural language applications, e.g., man-machine interface, automatic translation, text generation, *etc*. *Bag generation* (Brown, Cocke, *et al.*, 1990) is one of natural language generation methods. Given a sentence, we cut it up into words, place these words in a bag and try to recover the sentence from the bag. In corpus-based approach (Church and Mercer, 1993), a language model should be provided to measure the possible candidates. Markov Model (Kuhn and Mori, 1990) and word association model (Church and Hanks, 1990) are two famous models in language modeling. Markov Model has capabilities to keep the *linear precedence* relations in the context, so that it is useful to the application of bag generation. However, the parameters are tremendous in high order Markov Model. Word association model can capture the *long distance dependency* relations in the context under the postulation that the window size is the length of sentence. Thus, it is useful to the applications such as lexical selection. This paper will propose an *Approximate Markov Model*, which has merits of these two models.

## 2. Approximate Markov Model

Let $S=<*, w_1, w_2, ..., w_m, *>$ be an arrangement in bag generation. The star symbol marks the beginning ($w_0$) and the ending ($w_{m+1}$) of the sentence. The probability of $S$ in trigram Markov Model is measured as follows:

$$P(S) = P(<*, w_1, w_2, ..., w_m, *>)$$
$$\cong P(*) * P(w_1|*) * \prod_{i=0}^{m-1} P(w_{i+2}|w_i^{i+1})$$

$$= \frac{\prod_{i=0}^{m-1} P(w_i, w_{i+1}, w_{i+2})}{\prod_{i=1}^{m-1} P(w_i, w_{i+1})}$$

This formula utilizes trigram training table (numerator part) and bigram training table (denominator part) to compute the probability of an arrangement. It can be approximated by the following formula:

$$\frac{\prod_{i=0}^{m-1} \mathrm{Min}(P(w_i, w_{i+1}, 1), P(w_{i+1}, w_{i+2}, 1), P(w_i, w_{i+2}, 2))}{\prod_{i=1}^{m-1} P(w_i, w_{i+1}, 1)} \tag{1}$$

where   Min denotes a minimal function,
$P(w_i, w_j, j\text{-}i)$ is the probability of a directed word pair $(w_i, w_j)$ whose distance is j-i, e.g., $(w_i, w_{i+1}, 1)$ denotes $w_i$ is followed by $w_{i+1}$.

By the notation of directed word pair with distance, the statement "$w_{i+2}$ follows $w_{i+1}$ and $w_{i+1}$ follows $w_i$" (hereafter, $<w_i, w_{i+1}, w_{i+2}>$) can be represented as $(w_i, w_{i+1}, 1)$, $(w_{i+1}, w_{i+2}, 1)$ and $(w_i, w_{i+2}, 2)$. Consider the following figure. Assume parts (i), (ii) and (iii) correspond to the probabilities of $(w_i, w_{i+1}, 1)$, $(w_{i+1}, w_{i+2}, 1)$ and $(w_i, w_{i+2}, 2)$, respectively. In this way, part (iv) denotes the probability of $<w_i, w_{i+1}, w_{i+2}>$. From this figure, we know $P(w_i, w_{i+1}, w_{i+2}) \leq P(w_i, w_{i+1}, 1)$, $P(w_i, w_{i+1}, w_{i+2}) \leq P(w_{i+1}, w_{i+2}, 1)$ and $P(w_i, w_{i+1}, w_{i+2}) \leq P(w_i, w_{i+2}, 2)$. Thus, the minimum of $P(w_i, w_{i+1}, 1)$, $P(w_{i+1}, w_{i+2}, 1)$ and $P(w_i, w_{i+2}, 2)$ can be used to approximate $P(w_i, w_{i+1}, w_{i+2})$.

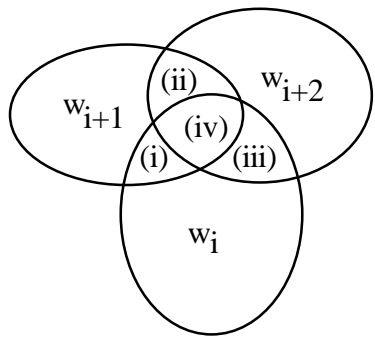

The model formulated by (1) is called *Approximate trigram Markov Model*. Similarly, the following n-gram Markov Model:

$$\begin{aligned} P(S) &= P(<*, w_1, w_2, ..., w_m, *>) \\ &\cong P(*)*P(w_1|*)*P(w_2|*, w_1)*...*P(w_{n\text{-}2}|w_0^{n-3})* \prod_{i=0}^{m-n+2} P(w_{i+n\text{-}1}|w_i^{i+n-2}) \end{aligned}$$

$$= \frac{\prod_{i=0}^{m-n+2} P(w_i^{i+n-1})}{\prod_{i=1}^{m-n+2} P(w_i^{i+n-2})}$$

can be approximated by:

$$\frac{\prod_{k=0}^{m-n+2} \text{Min}_{i,j(k \leq i < j \leq n+k-1)} P(w_i, w_j, j-i)}{\prod_{k=1}^{m-n+2} \text{Min}_{i,j(k \leq i < j \leq n+k-2)} P(w_i, w_j, j-i)} \quad (2)$$

Formula (2) denotes *Approximate n-gram Markov Model*. Assume the vocabulary size is V, and the average sentence length is L. The number of parameters of Approximate Markov Model is always $O((L-1)*V^2)$ no matter which order it has. Markov bigram and trigram Model have $O(V^2)$ and $O(V^3)$ parameters, respectively. The number of parameters multiplies by V when the order increases by one. Thus, Approximate Markov Model can be used to enlarge the window size, when the parameter issue is considered.

## 3. Bag Generation Algorithm

The bag generation algorithm under (Approximate) n-gram Markov Model is shown below.

---

**insert** *starting node* **into** *queue*
**while not** empty *queue* **do**
**begin**
    **initialize** an empty *list*
    **repeat**
        **remove** a *node* **from** *queue*, **and assign** it **to** *current node*
        **if** *current node* ≠ *final node* **then**
        **begin**
            **expand** *current node* **and**
            **merge** to the *list* if any two paths satisfy all of the following conditions:
            (1) the path length should be longer than n-1.
            (2) the lengths of these two paths should be equal.
            (3) the last n-1 nodes on these two paths should be equal.
            (4) these two paths should cover the same words.
        **end**
        **else merge** to the *list*
    **until** empty *queue*
    **if** *current node* ≠ *final node* **then assign** *list* **to** *queue*
**end**
**generate** the result from *list*, **and check** whether it is error or not.

The merge operation keeps the path with higher probability, and discards the path with lower probability. The four conditions in the above algorithm should be met if dynamic programming technique is used. The following proposition clarifies this point for Markov Model. Approximate Markov Model has the similar proof.

**Proposition.** The merge operation should obey the following conditions, if n-gram Markov Model is adopted:
  (1) The path length should be longer than n-1.
  (2) The lengths of these two paths should be equal.
  (3) The last n-1 nodes on these two paths should be equal.
  (4) These two paths should cover the same words.

*Proof*:

The first two are the basic definitions for n-gram Markov Model. In this model, the system will use the last n-1 words to predict the probability of the current word. Let the probabilities of two paths $H_1$ and $H_2$ be $P(H_1)$ and $P(H_2)$, and $P(H_1) > P(H_2)$. When the next word $w_m$ ($m \geq n-1$) is read, their probabilities become:

  $P(H_1) * P(w_m | w_{1(m-n+1)}, ..., w_{1(m-1)})$ and
  $P(H_2) * P(w_m | w_{2(m-n+1)}, ..., w_{2(m-1)})$, respectively.

If the last n-1 words are the same, i.e., $w_{1(m-n+1)} = w_{2(m-n+1)}, ..., w_{1(m-1)} = w_{2(m-1)}$, then the former is still larger than the later. However, if the last n-1 words on these two paths are not the same, then the former may be smaller than the latter. Thus, merging may introduce the error results.

In fact, the first three conditions are enough for the other Markov-based applications such as phone-to-text transcription, *etc*. However, there is a problem in bag generation application, if we do not obey the last condition either. Consider a general case. Let the two paths $H_1$ and $H_2$ have the following forms:

  $H_1$: $w_{10}, w_{11}, ..., w_{1(m-n)}, w_{(m-n+1)}, ..., w_{(m-1)}$ and
  $H_2$: $w_{20}, w_{21}, ..., w_{2(m-n)}, w_{(m-n+1)}, ..., w_{(m-1)}$.

If $\{w_{10}, w_{11}, ..., w_{1(m-n)}\}$ is not equal to $\{w_{20}, w_{21}, ..., w_{2(m-n)}\}$, there must exist some $w_{1i}$ and $w_{2j}$ such that $w_{1i} \neq w_{2j}$. If $P(H_1) > P(H_2)$, then the path involving $w_{1i}$, i.e.,

  $w_{20}, w_{21}, ..., w_{2(m-n)}, w_{(m-n+1)}, ..., w_{(m-1)}, w_{1i}$

will be neglected. This path may have higher probability, so that error occurs. ∎

The cost paid by the Approximate n-gram Markov Model is: each minimal value in the numerator part and denominator part of Formula (2) is derived from $n*(n-1)/2$ pairs and $(n-1)*(n-2)/2$ pairs, respectively. Consider the numerator part. For each tuple $<w_k, w_{k+1}, ..., w_{k+n-1}>$ ($0 \leq k \leq m-n+2$), its probability is determined by $P(w_i, w_j, j-i)$ ($k \leq i < j \leq n+k-1$). The complexity of an algorithm to select the minimum from $n*(n-1)/2$ pairs is $O(n^2)$. It is a terrible overhead. Here, a special data structure, i.e., a ring of n-1 elements, is adopted. Each element records the minimum of $k+n-1-i$ probabilities $P(w_i, w_{i+p}, p)$ ($1 \leq p \leq (n-1)-(i-k)$). The index i is ranged from k to $n+k-2$. The minimum of the $n*(n-1)/2$ pairs can be computed from these n-1 elements. When k is increased by one, i.e., the tuple $<w_{k+1}, w_{k+2}, ..., w_{k+n}>$ is inspected, only these (n-1) elements are considered instead of $n*(n-1)/2$ pairs. In other words, the position in the ring for $P(w_k, w_{k+p}, p)$ ($1 \leq p \leq n-1$) is free, and is used to record $P(w_{k+n-1},$ $w_{k+n}$, 1). $P(w_{k+p}, w_{k+n}, n-p)(1 \leq p \leq n-2)$ are compared with the corresponding elements in the ring. This can be done in $O(n)$ time.

## 4. Experimental Results

BDC corpus, which is a Chinese segmented corpus, is adopted as the source of the training data. It includes 7010 sentences about 50000 words. For each sentence $S=<*, w_1, w_2, ..., w_m, *>$ in the training corpus, total $(m+1)*(m+2)/2$ directed word association pairs, which are of the form $(w_i, w_j, j-i)$(where $0 \leq i < j \leq m+1$), are generated. The experimental results (distribution of error sentences) of bag generation by using Markov Model and Approximate Markov Model are shown in the following table. M*i* and AM*i* denote *i*-gram Markov Model and Approximate Markov Model, respectively.

| sentence length | total test sentences | Markov Model | | | | Approximate Markov Model | | | | |
|---|---|---|---|---|---|---|---|---|---|---|
| | | M2 | M3 | M4 | M5 | AM2 | AM3 | AM4 | AM5 | AMn |
| 1 | 6 | 0 | 0 | 0 | 0 | 0 | 0 | 0 | 0 | 0 |
| 2 | 34 | 0 | 0 | 0 | 0 | 0 | 0 | 0 | 0 | 0 |
| 3 | 121 | 0 | 0 | 0 | 0 | 0 | 0 | 0 | 0 | 0 |
| 4 | 213 | 1 | 0 | 0 | 0 | 1 | 0 | 0 | 0 | 0 |
| 5 | 297 | 0 | 0 | 0 | 0 | 0 | 0 | 0 | 0 | 0 |
| 6 | 329 | 3 | 0 | 0 | 0 | 3 | 0 | 0 | 0 | 0 |
| 7 | 234 | 4 | 0 | 0 | 0 | 4 | 2 | 1 | 0 | 0 |
| 8 | 216 | 11 | 0 | 0 | 0 | 11 | 1 | 1 | 0 | 0 |
| 9 | 183 | 6 | 0 | 0 | 0 | 6 | 0 | 0 | 0 | 0 |
| 10 | 170 | 8 | 0 | 0 | 0 | 8 | 0 | 0 | 0 | 0 |
| 11 | 129 | 11 | 0 | 0 | 0 | 11 | 0 | 0 | 0 | 0 |
| 12 | 68 | 13 | 0 | 0 | 0 | 13 | 1 | 1 | 0 | 0 |
| total | 2000 | 57 | 0 | 0 | 0 | 57 | 4 | 3 | 0 | 0 |

It is trivial that AM2 is equal to M2. The other results demonstrate that the power of approximate Markov Model is close to that of Markov Model.

## 5. Concluding Remarks

This paper proposes a directed word association model with distance to approximate Markov Model. It can increase the order of language model, and keep the number of parameters unchanged. The experimental results show that the performance of Approximate Markov Model and Markov Model is very close. Besides, the training knowledge for bag generation can be also applied to lexical selection. The co-occurrence of a word pair can be computed easily by sum of the related directed word association pairs. The uniform knowledge facilitates statistics-based machine translation design.